\documentclass[10pt,twocolumn,letterpaper]{article}

\usepackage{iccv}
\usepackage{times}
\usepackage{epsfig}
\usepackage{graphicx}
\usepackage{amsmath}
\usepackage{amssymb}

\usepackage{booktabs}
\usepackage{multirow}
\usepackage{bbm}
\usepackage{color}
\usepackage{xcolor}
\usepackage{mathtools}

\usepackage[breaklinks=true,bookmarks=false,colorlinks=true]{hyperref}

\iccvfinalcopy 

\def\iccvPaperID{1345} 

\ificcvfinal\fi

\begin{document}

\title{Crossover Learning for Fast Online Video Instance Segmentation}

\author{Shusheng Yang$^{1,2}$\thanks{Equal contributions. This work was done while Shusheng Yang was interning at Applied Research Center (ARC), Tencent PCG.}, \ \ Yuxin Fang$^{1*}$, \ \ Xinggang Wang$^{1}$\thanks{Corresponding author, E-mail: {\tt xgwang@hust.edu.cn.}}, \ \ Yu Li$^{2}$, \\
\vspace{0.15cm}
Chen Fang$^{3}$, \ \ Ying Shan$^{2}$, \ \ Bin Feng$^{1}$, \ \ Wenyu Liu$^{1}$ \\

$^1$School of EIC, Huazhong University of Science \& Technology \\
$^2$Applied Research Center (ARC), Tencent PCG \ \ $^3$Tencent\\
}

\maketitle
\ificcvfinal\thispagestyle{empty}\fi

\begin{abstract}
Modeling temporal visual context across frames is critical for video instance segmentation (VIS) and other video understanding tasks. In this paper, we propose a fast online VIS model named CrossVIS. For temporal information modeling in VIS, we present a novel crossover learning scheme that uses the instance feature in the current frame to pixel-wisely localize the same instance in other frames. Different from previous schemes, crossover learning does not require any additional network parameters for feature enhancement. By integrating with the instance segmentation loss, crossover learning enables efficient cross-frame instance-to-pixel relation learning and brings cost-free improvement during inference. Besides, a global balanced instance embedding branch is proposed for more accurate and more stable online instance association. We conduct extensive experiments on three challenging VIS benchmarks, \ie, YouTube-VIS-2019, OVIS, and YouTube-VIS-2021 to evaluate our methods. To our knowledge, CrossVIS achieves state-of-the-art performance among all online VIS methods and shows a decent trade-off between latency and accuracy. Code will be available to facilitate future research.
\end{abstract}

\section{Introduction}
	
	Video instance segmentation (VIS) \cite{VIS} is an emerging task in computer vision that aims to perform per-pixel labeling of instances within video sequences. 
	This task provides a natural understanding of the video scenes.
	Therefore achieving accurate, robust, and fast video instance segmentation in real-world scenarios will greatly stimulate the development of computer vision applications, \eg, autonomous driving, video surveillance, and video editing.

	Recently, signiﬁcant progress has been witnessed in still-image object detection and instance segmentation. However, extending these methods to VIS remains a challenging work.  Similar to other video-based recognition tasks, such as video object segmentation (VOS) \cite{DAVIS16, DAVIS17, GCMVOS}, video object detection (VOD) \cite{ImageNetVID, DSFNet} and multi-object tracking (MOT) \cite{ MOTChallenge, KITTI, MOTS, FairMOT}, continuous video sequences always bring great challenges, \eg, a huge number of frames required to be fast recognized, heavy occlusion, object disappearing and unconventional object-to-camera poses \cite{feichtenhofer2017detect}.

    \begin{figure}[tp]
    \centering
    \includegraphics[width=1.0\columnwidth]{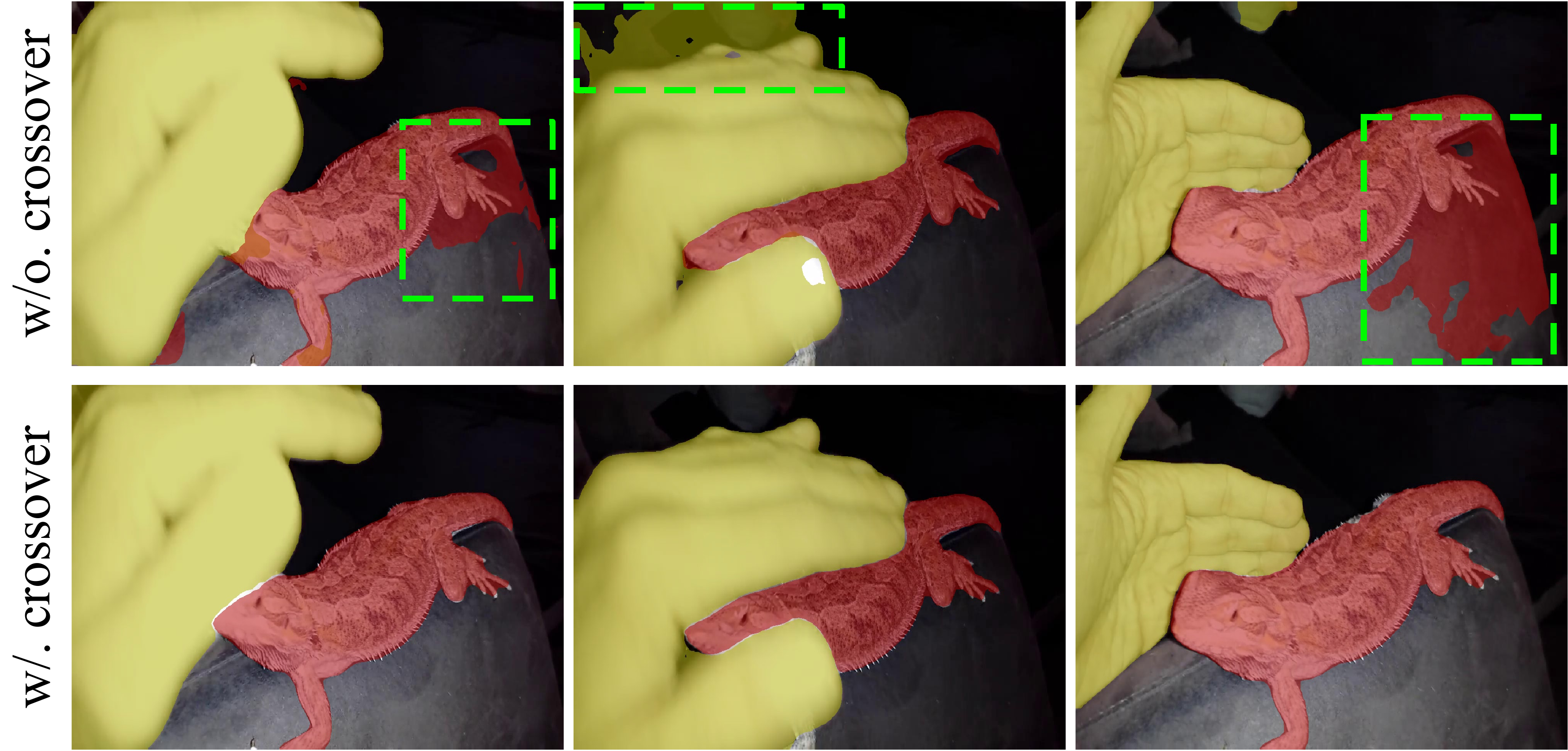}
    \caption{CrossVIS can predict more accurate video instance segmentation results (bottom row) compared with the baseline model without crossover learning (top row). More qualitative comparisons on YouTube-VIS-2019 $\mathtt{val}$ set are available in the Appendix.}
    \label{fig: Teaser}
\end{figure}

    To conquer these challenges and obtain better performance on these video understanding tasks (VIS, VOS, VOD, and MOT), fully utilizing the temporal information among video frames is critical. Previous deep learning based methods on this topic are in four folds. $(1)$ Pixel-level feature aggregation enhances pixels feature of the current frame using other frames, \eg, STM-VOS \cite{STM} and STEm-Seg \cite{STEm-Seg} aggregates pixel-level space-time feature based on Non-local network \cite{wang2018non} and 3D convolution, respectively. $(2)$ Instance-level feature aggregation enhances region, proposal or instance features across frames, \eg, MaskProp \cite{MaskPropagation} propagates instance features using deformable convolution \cite{DCNv1} for VIS and SELSA \cite{wu2019sequence} fuses instance features using spectral clustering for VOD. $(3)$ Associating instances using metric learning, \eg, MaskTrack R-CNN \cite{VIS} introduces an association head based on Mask R-CNN \cite{MaskRCNN} and SipMask-VIS \cite{SipMask} adds an adjunctive association head based on FCOS \cite{FCOS}. $(4)$ Post-processing, \eg, Seq-NMS \cite{han2016seq} and ObjLink \cite{tang2020object} refine video object detection results based on dynamic programming and learnable object tubelet linking, respectively.

    In this paper, we propose a new scheme for temporal information modeling termed crossover learning. The basic idea is to use the instance feature in the current frame to pixel-wisely localize the same instance in other frames. Different from previous pixel/instance-level feature aggregation methods, crossover learning does not require additional network blocks for feature alignment and fusion. It obtains temporal information enhanced features without increasing inference computation cost. Different from metric learning based instance associating methods that require additional metric learning losses, crossover learning is integrated with the instance segmentation loss. Besides, it enables efficient many-to-many relation learning across frames, \ie, the instance pixel features are enforced to be close to the pixels that belong to the same instance and far from pixels that belong to other instances and background.
    Different from the post-processing methods, crossover learning is end-to-end optimizable with back-propagation.

	
	Since crossover learning is integrated with the instance segmentation loss, it is fully compatible with the other temporal information modeling strategies. 
	In this paper, we further improve the instance association strategy by introducing a global balanced instance embedding learning network branch. 
	Our main contributions are summarized as follows: 
	
	\begin{itemize}
	    \item We propose a novel crossover learning scheme that leverages the rich contextual information inherent in videos to strengthen the instance representation across video frames, and weaken the background and instance-irrelevant information in the meantime.
	   
	    \item We introduce a new global balanced instance embedding branch to tackle the association problem in video instance segmentation, which yields better and more stable results than previous pair-wise identity mapping approaches.
	    
	    \item We propose a fully convolutional online video instance segmentation model CrossVIS that achieves a promising result on three challenging VIS benchmarks, \ie, YouTube-VIS-2019, OVIS, and YouTube-VIS-2021. To our knowledge, CrossVIS achieves state-of-the-art performance among all online VIS methods and strikes a decent speed-accuracy trade-off.
	   
	\end{itemize}

\section{Related Work}

\noindent \textbf{Still-image Instance Segmentation.} \
 	Instance segmentation is the task of detecting and segmenting each distinct object of interest in a given image. Many prior works \cite{MaskRCNN, MaskLab, InstanceAwareSS, MSRCNN, HTC, TensorMask, YOLACT, SOLO, CondInst, CenterMask, SipMask, PolyTransform} contribute a lot to the rapid developments in this field. 
 	Mask R-CNN \cite{MaskRCNN} adapts Faster R-CNN \cite{FasterRCNN} with a parallel mask head to predict instance masks, and leads the two-stage fashion for a long period of time. 
 	\cite{MSRCNN, HTC, BPMask} promote Mask R-CNN and achieve better instance segmentation results. 
 	The success of these two-stage models partially is due to the feature alignment operation, \textit{i.e.}, RoIPool \cite{SpatialPyramidPooling, FastRCNN} and RoIAlign \cite{MaskRCNN}.
 	Recently, instance segmentation methods based on one-stage frameworks without explicit feature alignment operation begin to emerge \cite{YOLACT, YOLACTpp, BlendMask, PolarMask, SOLO, SOLOv2}.
 	As a representative, the fully convolutional CondInst \cite{CondInst} outperforms several state-of-the-art methods on the COCO dataset \cite{COCO}, which dynamically generates filters for mask head conditioned on instances. 
 	We build our framework on top of \cite{CondInst} and extend it to the VIS task.

\noindent
\textbf{Video Instance Segmentation (VIS).} \ 
 	VIS requires classifying, segmenting, and tracking visual instances over all frames in a given video.
 	With the introduction of YouTube-VIS-2019 dataset \cite{VIS}, tremendous progresses \cite{VIS-WS1, VIS-WS2, VIS-WS3, VIS-WS4} have been made in tackling this challenging task. 
 	As a representative method, MaskTrack R-CNN \cite{VIS} extends the two-stage instance segmentation model Mask R-CNN with a pair-wise identity branch to solve the instance association sub-task in VIS.
 	SipMask-VIS \cite{SipMask} follows the similar pipeline based on the one-stage FCOS \cite{FCOS} and YOLACT \cite{YOLACT, YOLACTpp} frameworks.
 	\cite{VISWinner} separates all sub-tasks in VIS problem and designs specific networks for each of them, all networks are trained independently and combined during inference to generate the final predictions.
 	MaskProp \cite{MaskPropagation} introduces a novel mask propagation branch on the multi-stage framework \cite{HTC} that propagates instance masks from one frame to another. 
 	As an offline method, MaskProp achieves accurate predictions but suffers from high latency. 
 	\cite{VAE-VIS} introduces a modified variational auto-encoder to solve the VIS task. 
 	STEm-Seg \cite{STEm-Seg} treats the video clip as 3D spatial-temporal volume and segments objects in a bottom-up fashion. 
 	\cite{VIS_RGNN} adopts recurrent graph neural networks for VIS task.
 	CompFeat \cite{CompFeat} refines features at both frame-level and object-level with temporal and spatial context information.
 	VisTR \cite{VisTR} naturally adopts DETR \cite{DETR} for VIS task in a query-based end-to-end fashion.
 	
 	Recently, more challenging benchmarks such as OVIS \cite{OVIS} and YouTube-VIS-2021 \cite{YouTube-VIS-2021} are proposed to further promote the advancement of this field.
 	CrossVIS is evaluated on three VIS benchmarks and shows competitive performances.
 	We hope CrossVIS can serve as a strong baseline to facilitate future research.

\begin{figure*}
  \centering
  \includegraphics[width=1.6\columnwidth]{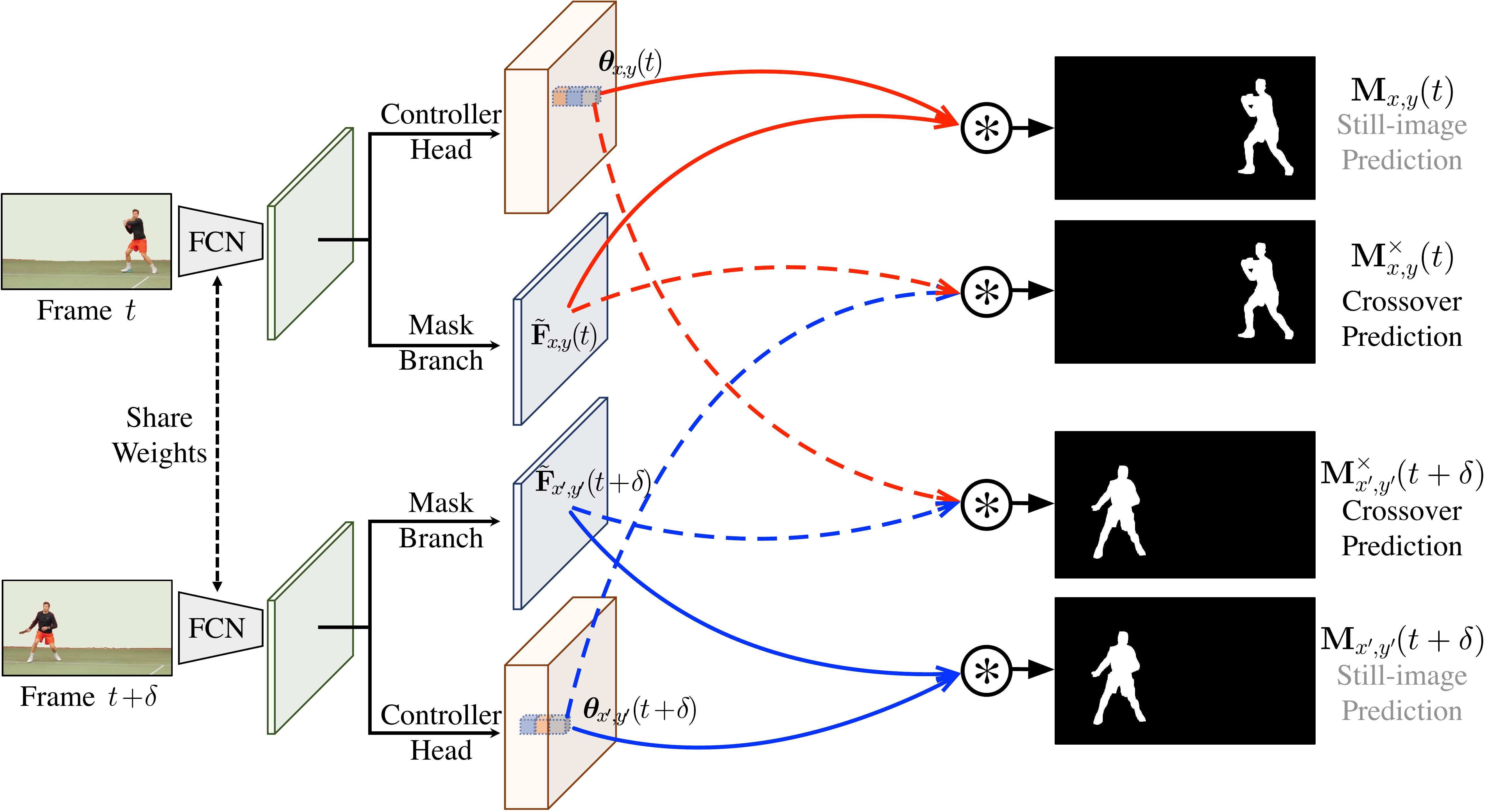}
  \caption{Overview of CrossVIS in the training phase. 
  Two frames at time $t$ and $t + \delta$ are fed into an fully convolutional network (FCN) to generate dynamic filters $\boldsymbol{\theta}_{x, y}(t)$ \& $\boldsymbol{\theta}_{x^\prime, y^\prime}(t + \delta)$ and mask feature maps $\tilde{\mathbf{F}}_{x, y}(t)$ \& $\tilde{\mathbf{F}}_{x^\prime, y^\prime}(t + \delta)$. 
  Red lines indicate the dynamic filters and mask feature maps in frame $t$, blue lines indicate the same in frame $t + \delta$. Solid lines indicate the still-image prediction process, dotted lines indicate the proposed crossover learning scheme. The four ``\mbox{\large $\circledast$}" from top to bottom in the figure correspond to the mask generation process formulated in Eq.~\eqref{eq: M_xy(t)}, Eq.~\eqref{eq: M_xy(t)^crossover}, Eq.~\eqref{eq: M_xprimeyprime(t+delta)^crossover}, and Eq.~\eqref{eq: M_xprimeyprime(t+delta)}, respectively. Classification, localization as well as global balanced instance embedding branches are omitted in the figure for clarification.}
  \label{fig: CrossVIS}
\end{figure*}

\section{Method}

    Our goal is to leverage the rich contextual information across different video frames for a more robust instance representation in video instance segmentation (VIS). 
    To this end, we take inspiration from \cite{DFN, CondInst, SOLOv2} and propose CrossVIS (see Fig.~\ref{fig: CrossVIS}) that consists of two key components tailor-made for VIS task:
    $(1)$ the crossover learning scheme for more accurate video-based instance representation learning, and 
    $(2)$ the global balanced instance embedding branch for better online instance association.

\subsection{Mask Generation for Still-image}
\label{sec: still-img instance seg}
    For still-image instance segmentation, we  leverage the dynamic conditional convolutions \cite{DFN, CondInst}.
    Specifically, our method generates the instance mask $\mathbf{M}_{x, y}$ at location $(x, y)$ by convolving an instance-agnostic feature map $\tilde{\mathbf{F}}_{x, y}$ from the mask branch and a set of instance-specific dynamic filters $\boldsymbol{\theta}_{x, y}$ produced by the controller head. 
    Formally:
	\begin{equation}
	    \tilde{\mathbf{F}}_{x, y} = \mathrm{Concat}\Big(\mathbf{F}_{mask}; \mathbf{O}_{x, y}\Big),
	\end{equation}
	\begin{equation}
		\mathbf{M}_{x, y} = \mathrm{MaskHead}\Big(\tilde{\mathbf{F}}_{x, y}; \boldsymbol{\theta}_{x, y}\Big),
	\end{equation}
    where $\tilde{\mathbf{F}}_{x, y}$ is the combination of mask feature map $\mathbf{F}_{mask}$ and relative coordinates $\mathbf{O}_{x, y}$. $\mathbf{F}_{mask}$ is produced via the mask branch attached on FPN \cite{FPN} $\{P_3, P_4, P_5\}$ level features.
    The relative coordinates $\mathbf{O}_{x, y}$ provide a strong localization cue for predicting the instance mask.
    The $\mathrm{MaskHead}$ consists of $3$ conv-layers with dynamic filters $\boldsymbol{\theta}_{x, y}$ conditioned on the instance located at $(x, y)$ as convolution kernels.
    The last layer has $1$ output channel and uses $\mathrm{sigmoid}$ function for instance mask predictions.
	
\subsection{Crossover Learning}
\label{sec: crossover}
    \noindent \textbf{Intuition of Crossover Learning.} \ 
    Still-image instance segmentation needs two types of information \cite{CondInst}:
    $(1)$ appearance information to categorize objects, which is given by the dynamic filter $\boldsymbol{\theta}_{x, y}$ in our model; 
    and $(2)$ location information to distinguish multiple objects belonging to the same category, which is represented by the relative coordinates $\mathbf{O}_{x, y}$. 
    In the aforementioned still-image instance segmentation model (see Sec.~\ref{sec: still-img instance seg}), for each instance, we have a one-to-one correspondence between the appearance information and location information: given a $\boldsymbol{\theta}_{x, y}$, there exists one and only one $\mathbf{O}_{x, y}$ as the corresponding location information belonging to the same instance.
    Meanwhile, the connection between different instances is \textit{isolated}.
    
    However, in terms of the VIS task, given a sampled frame-pair from one video, the same instance may appear in different locations of two different sampled frames.
    Therefore it is possible to use the appearance information from one sampled frame to represent the \textit{same} instance in two \textit{different} sampled frames, guided by \textit{different} location information.
    We can utilize the appearance information $\boldsymbol{\theta}_{x, y}(t)$ from one sampled frame $t$ to incorporate the location information $\mathbf{O}_{x, y}(t + \delta)$ of the \textit{same instance in another sampled frame} $t + \delta$. 
    By this kind of across frame mapping, we expect the learned instance appearance information can be enhanced and more robust, meanwhile, the background and instance-irrelevant information is weakened.

	\noindent \textbf{Formulation of Crossover Learning.} \  
	Specifically, for a given video, we denote a detected instance $i$ at time $t$ (or frame $t$) as:
	\begin{equation}
    	\mathcal{I}_{i}(t) = (c_{i}(t), \boldsymbol{\theta}_{x,y}(t), \mathbf{e}_{i}(t)),
	\end{equation}
	where $c_{i}(t)$ is the instance category, $\boldsymbol{\theta}_{x,y}(t)$ is the dynamic filter for $\mathrm{MaskHead}$, and $\mathbf{e}_{i}(t)$ is the instance embedding for online association.
	Without loss of generality, we assume that an instance $\mathcal{I}_{i}$ exists in both frame $t$ (denoted as $\mathcal{I}_{i}(t)$) as well as frame $t + \delta$ (denoted as $\mathcal{I}_{i}(t + \delta)$). 
	
	Within each frame, following the setup and notation in Sec.~\ref{sec: still-img instance seg}, at time $t$, the instance mask of $\mathcal{I}_{i}(t)$ located at $(x, y)$ can be represented as:
	\begin{equation}
	\label{eq: M_xy(t)}
		\mathbf{M}_{x, y}(t) = \mathrm{MaskHead}\Big(\tilde{\mathbf{F}}_{x, y}(t) ; \boldsymbol{\theta}_{x, y}(t)\Big).
	\end{equation}
	At time $t + \delta$, the instance move from location $(x, y)$ to location $(x^{\prime}, y^{\prime})$.
	So the instance mask of $\mathcal{I}_{i}(t + \delta)$ can be represented as:
	\begin{equation}
    \label{eq: M_xprimeyprime(t+delta)}
		\mathbf{M}_{x^{\prime}, y^{\prime}}(t + \delta) = \mathrm{MaskHead}\Big(\tilde{\mathbf{F}}_{x^{\prime}, y^{\prime}}(t + \delta) ; \boldsymbol{\theta}_{x^{\prime}, y^{\prime}}(t + \delta)\Big), 
	\end{equation}

    Our crossover learning scheme establishes a connection between the dynamic filter from one frame and the mask feature map from another frame. 
    Specifically, we expect the dynamic filter $\boldsymbol{\theta}_{x, y}(t)$ of $\mathcal{I}_{i}(t)$ can produce the mask of $\mathcal{I}_{i}(t + \delta)$ by convolving its mask feature map $\tilde{\mathbf{F}}_{x^{\prime}, y^{\prime}}(t + \delta)$: 
	\begin{equation}
	\label{eq: M_xprimeyprime(t+delta)^crossover}
		\mathbf{M}^{\boldsymbol{\times}}_{x^{\prime}, y^{\prime}}(t + \delta) =  \mathrm{MaskHead}\Big(\tilde{\mathbf{F}}_{x^{\prime}, y^{\prime}}(t + \delta) ; \boldsymbol{\theta}_{x, y}(t)\Big), 
	\end{equation}
	where $\mathbf{M}^{\boldsymbol{\times}}$ with a superscript ``$\times$" denotes the instance mask produced by crossover learning. Similarly, we expect the dynamic filter $\boldsymbol{\theta}_{x^{\prime}, y^{\prime}}(t + \delta)$ of $\mathcal{I}_{i}(t + \delta)$ can produce the mask of $\mathcal{I}_{i}(t)$ by convolving its mask feature map $\tilde{\mathbf{F}}_{x, y}(t)$: 
	\begin{equation}
	\label{eq: M_xy(t)^crossover}
		\mathbf{M}^{\boldsymbol{\times}}_{x, y}(t) = \mathrm{MaskHead}\Big(\tilde{\mathbf{F}}_{x, y}(t) ; \boldsymbol{\theta}_{x^{\prime}, y^{\prime}}(t + \delta)\Big).
	\end{equation}
	Following \cite{CondInst}, during training, the predicted instance masks $\mathbf{M}_{x, y}(t)$, $\mathbf{M}_{x^{\prime}, y^{\prime}}(t + \delta)$, $\mathbf{M}^{\boldsymbol{\times}}_{x^{\prime}, y^{\prime}}(t + \delta) $ and $\mathbf{M}^{\boldsymbol{\times}}_{x, y}(t)$ are all optimized by the dice loss \cite{DiceLoss}:
    \begin{equation}
    \label{eq: dice loss}
        \mathcal{L}_{dice}\left(\mathbf{M}, \mathbf{M}^{\ast}\right) 
        = 1 - \frac{2 \sum_{i}^{HW} \mathbf{M}_{i} \mathbf{M}^{\ast}_{i}}{\sum_{i}^{HW}\left(\mathbf{M}_{i}\right)^{2}+\sum_{i}^{HW}\left(\mathbf{M}^{*}_{i}\right)^{2}}
    \end{equation}
    where $\mathbf{M}$ is the predicted mask and $\mathbf{M}^*$ is the ground truth mask, $i$ denotes the $i^{th}$ pixel.
	During inference, the instance mask generation process keeps the same as \cite{CondInst}, with no crossover involved.
	
	\noindent \textbf{Advantages of Crossover Learning.} \
	For a given instance $\mathcal{I}_{i}(t)$, its appearance information $\boldsymbol{\theta}_{x, y}(t)$ can learn two kinds of representations: a within-frame one in frame $t$, and an across-frame one in frame $t + \delta$. 
    At time $t + \delta$, the instance $\mathcal{I}_i(t + \delta)$ may have a different appearance and be in a different context compared with the same instance $\mathcal{I}_i(t)$ at time $t$. Meanwhile, the background may also changed.
    The crossover learning enables dynamic filter $\boldsymbol{\theta}_{x, y}(t)$ to identify the same instance representation at both time $t$ and $t + \delta$, regardless of the background and instance-irrelevant information.
	In this way, we can largely overcome the appearance inconsistency as well as background clutters problems in videos, leveraging the rich contextual information across video frames to get a more accurate and robust instance representation.

\begin{figure}
    \centering
    \includegraphics[width=1\columnwidth]{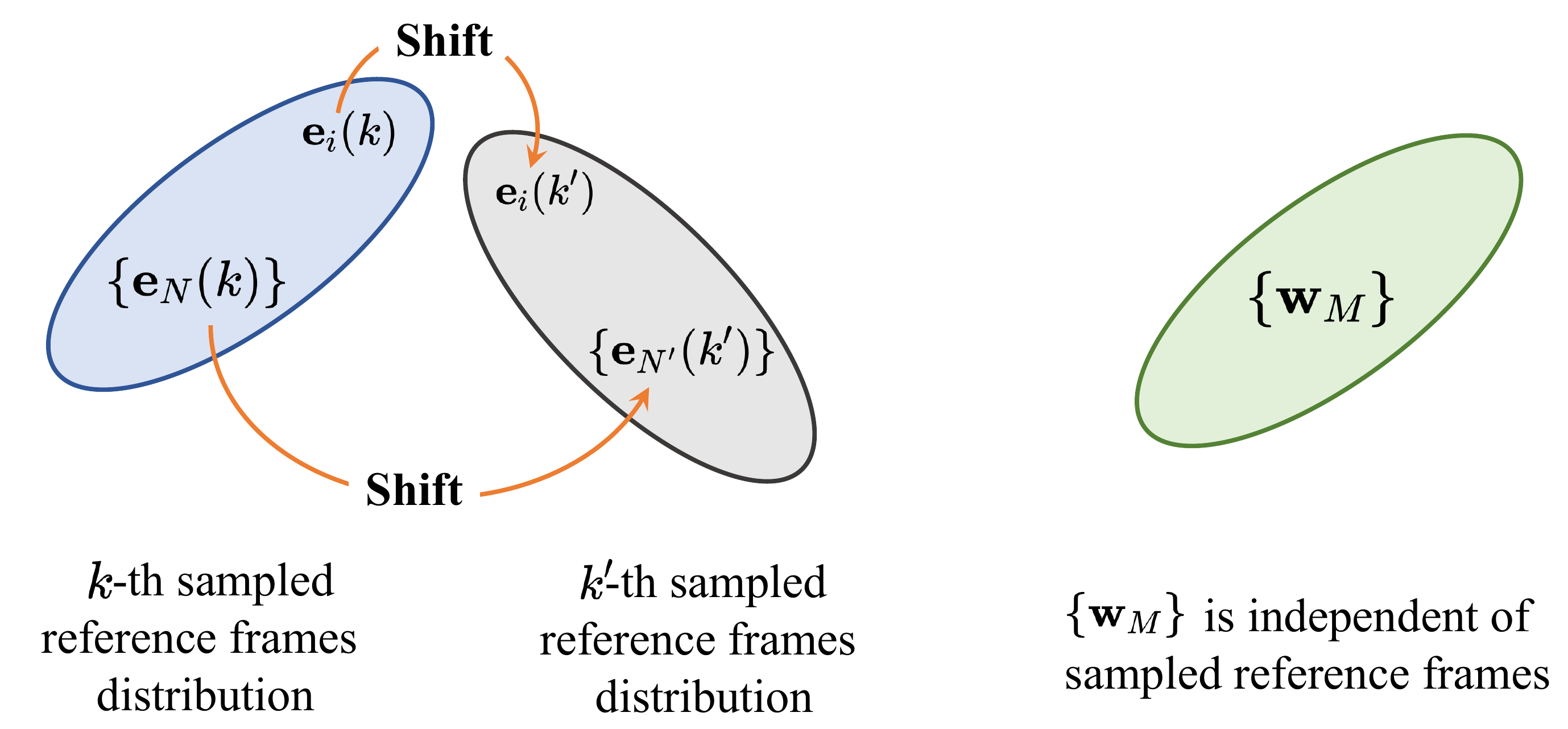}
    \caption{An illustration of pair-wise local embeddings (Fig.~\ref{fig: global balanced embedding}, left) used in \cite{VIS, SipMask}, and the proposed instance proxies (Fig.~\ref{fig: global balanced embedding}, right). 
    For pair-wise local embeddings, $\{\mathbf{e}_N(k)\}$ is the set of all $N$ instance embeddings from $k$-th sampled reference frames, while in the $k^\prime$-th sampling, the sampled instance identities will change to $\{\mathbf{e}_{N^\prime}(k^\prime)\}$, causing a distribution shift.
    Even if the same instance $\mathcal{I}_i$ is happened to be sampled in both $k$-th and $k^\prime$-th samplings, the corresponding embedding $\{\mathbf{e}_i(k)\}$ may also shift to $\{\mathbf{e}_i(k^\prime)\}$ due to occlusion, changing of background and scale variation, \etc. 
    In contrast, $\{\mathbf{w}_M\}$ is a set of learnable instance-wise weights of the model and independent of sampled reference frames. Therefore $\{\mathbf{w}_M\}$ produces a global, definite convergence status for instance embeddings.
}
    \label{fig: global balanced embedding}
\end{figure}

\subsection{Learning Global Balanced Embeddings for \\ Instance Association}
\label{sec: instance embeddings}
    Another crucial sub-task in VIS is the instance association, \ie, learning instance embeddings where instances of the same identity are close to each other in the feature space, while instances that belong to different identities are far apart.
    These embeddings are used for online inference.
    
    In previous tracking-by-detection VIS methods \cite{VIS, SipMask}, the instance embedding is trained in a \textit{pair-wise} and \textit{local} manner.
    Specifically, given a key frame at time $t + \delta$ and a reference frame at time $t$, assuming there is a detected candidate instance $\mathcal{I}_i$ in the key frame as the training sample, and $N$ already identified instances (given by ground truth label during training) in the reference frame as targets. Then, $\mathcal{I}_i$ can only be assigned to one of the $N$ identities if it is one of the already identified instances or a new identity if it is a new instance.
    The probability of assigning label $n$ to $\mathcal{I}_i$ is defined as:
    \begin{equation}
        \label{eq: pair-wise ce id loss 1}
        p_i(n) = 
        \begin{cases}
            \dfrac{\exp{(\mathbf{e}_{i}^{\top}\mathbf{e}_n})}{1 + \sum_{j = 1}^{N}
            \exp{(\mathbf{e}_{i}^{\top}\mathbf{e}_j})} & \mathrm{if \ } n \in [1, N],\\ 
            & \\
            \dfrac{1}{1 + \sum_{j = 1}^{N}
            \exp{(\mathbf{e}_{i}^{\top}\mathbf{e}_j})} & \mathrm{otherwise},
        \end{cases}
    \end{equation}
    where $\mathbf{e}_i$ and $\mathbf{e}_n$ denote the instance embedding of $\mathcal{I}_i$ from the key frame and $\mathcal{I}_n$ from the reference frame, respectively.
    $p_i(n)$ is optimized by cross-entropy loss:
    \begin{equation}
    \label{eq: pair-wise ce id loss 2}
        \mathcal{L}_{CE} = -\log(p_i(n)).
    \end{equation}

    However, this approach suffers from the following issues (see Fig~\ref{fig: global balanced embedding}, left) \cite{CircleLoss, QuasiDense}: 
    the feature space where $\mathbf{e}_i$ and $\{\mathbf{e}_N\} \coloneqq \{\mathbf{e}_1, \mathbf{e}_2, \dots, \mathbf{e}_N\}$ live in is defined by the sampled frames, and the decision boundary is closely related to the instance embeddings $\{\mathbf{e}_N\}$ from the reference frame.
    Therefore, the optimization and instance association processes highly depend on stochastic frame sampling, which probably lead to unstable learning and slow convergence. 
    We also observe a relatively large fluctuation in AP when using pair-wise embeddings (see $\sigma_{\text{AP}}$ in Tab.~\ref{tab: ablation for instance association embedding}).
    
    To remedy these problems and get a globally definite convergence status for instance embeddings, we train our model as a $M$-class classification problem where $M$ equals to the number of all different identities in the whole training set. 
    We then employ a set of learnable instance-wise weights $\{\mathbf{w}_M\} \coloneqq \{\mathbf{w}_1, \mathbf{w}_2, \dots, \mathbf{w}_M\}$ as proxies of instances (see Fig~\ref{fig: global balanced embedding}, right) to replace the embeddings of instances $\{\mathbf{e}_N\}$ defined by the sampled frame pair directly \cite{Metric_1, CloserFSL, JDE}. 
    In this way, the probability of assigning label $n$ to $\mathcal{I}_i$ is reformulated as:
    \begin{equation}
    \label{eq: global ce id loss 1}
        p_i(n) = 
            \dfrac{\exp{(\mathbf{e}_{i}^{\top}\mathbf{w}_n})}{\sum_{j = 1}^{M}
            \exp{(\mathbf{e}_{i}^{\top}\mathbf{w}_j})}.
    \end{equation}
    $p_i(n)$ is also optimized by cross-entropy loss: 
    \begin{equation}
    \label{eq: global ce id loss 2}
        \mathcal{L}_{CE} = -\log(p_i(n)).
    \end{equation}
    
    However, the $M$-class classification problem is hard to extend to large-scale datasets (\textit{e.g.}, $M = 3,774$ for the YouTube-VIS-2019 training set) as all the negative classes participate in the loss computation, resulting in a large pos-neg samples imbalance issue.
    Moreover, the large gradient produced by these negative samples from the instance embedding branch dominates the learning process\footnote{For the classification sub-task, the large amount of easy negative samples can be handled by focal loss \cite{FocalLoss}. For regression and segmentation sub-tasks, only positive samples participate in training.}, which can negatively affect the optimization of all sub-tasks.
    To remedy these problems, we adopt focal loss \cite{FocalLoss} as the objective for our global instance embedding to balance the pos-neg samples as well as the learning of each sub-task:
    \begin{equation}
	\label{eq: global focal id loss 1}
        p_i(n) = 
        \begin{cases}
             \hspace{1.7em} \sigma(\mathbf{e}_i^{\top} \mathbf{w}_n) & \mathrm{if \ } \mathcal{I}_i = \mathcal{I}_n ,\\ 
            
            1 - \sigma(\mathbf{e}_i^{\top} \mathbf{w}_n) & \mathrm{otherwise},
        \end{cases}
    \end{equation}
    \begin{equation}
	\label{eq: global focal id loss 2}
        \mathcal{L}_{id} = \mathcal{L}_{Focal} = - \alpha_t (1 - p_i(n))^{\gamma}\log(p_i(n)),
    \end{equation}
    where $\sigma(\cdot)$ is the sigmoid function, $\alpha_t$ and $\gamma$ follow the definition in \cite{FocalLoss}.
    $\mathcal{I}_i = \mathcal{I}_n$ means the two instances belong to the same identity.
    $\mathbf{e}_i$ is generated by the proposed global balanced instance embedding branch which shares a common structure as the classification branches of \cite{CondInst}.

\subsection{Training and Online Inference}
\label{sec: training and online inference }
	We jointly train detection, segmentation, crossover learning and instance association tasks in an end-to-end manner. 
	The multi-task loss for each sample is:
	\begin{equation}
		\mathcal{L}
		= \mathcal{L}_{det}
		+ \mathcal{L}_{seg}
		+ \mathcal{L}_{cross} 
		+ \mathcal{L}_{id}.
	\end{equation}
	$\mathcal{L}_{det}$ and $\mathcal{L}_{seg}$ denote the object detection loss and still-image instance segmentation loss in \cite{CondInst}.
	$\mathcal{L}_{cross}$ denotes the crossover learning loss:
	\begin{equation}
	\begin{split}
	    \mathcal{L}_{cross} 
	    & = \mathcal{L}_{dice}(\mathbf{M}_{x, y}^{\times}(t), \mathbf{M}_{x, y}^{*}(t)) \\
	    & + \mathcal{L}_{dice}(\mathbf{M}_{x^{\prime}, y^{\prime}}^{\times}(t+\delta), \mathbf{M}_{x^{\prime}, y^{\prime}}^{*}(t+\delta)), 
	\end{split}
	\end{equation}
	where $\mathcal{L}_{dice}$ is formulated in Eq.~\eqref{eq: dice loss}.
	$\mathcal{L}_{id}$ denotes the instance embedding loss defined in Eq.~\eqref{eq: global focal id loss 1} \& Eq.~\eqref{eq: global focal id loss 2}.
	
	During inference, the testing video is processed by CrossVIS frame by frame in an online fashion.
    We follow the inference procedure described in \cite{VIS, CompFeat}.

\begin{table*}
 \setlength{\tabcolsep}{9.2 pt}
 \begin{small}
 \begin{tabular}{l|l|c|c|c|ccccc}
 \hline
 
 \hline
  Methods 
  & Backbone
  & Aug. 
  & Type
  & FPS
  & AP 
  & AP$_{\mathtt{50}}$ 
  & AP$_{\mathtt{75}}$ 
  & AR$_{\mathtt{1}}$ 
  & AR$_{\mathtt{10}}$ \\
  \hline
  \hline
  IoUTracker+ \cite{VIS} & ResNet-$50$ & & Online & - & $23.6$ & $39.2$ & $25.5$ & $26.2$ & $30.9$ \\
  OSMN \cite{OSMN} & ResNet-$50$ & & Online & - & $27.5$ & $45.1$ & $29.1$ & $28.6$ & $33.1$ \\
  DeepSORT \cite{DeepSort} & ResNet-$50$ & & Online & - & $26.1$ & $42.9$ & $26.1$ & $27.8$ & $31.3$ \\
  FEELVOS \cite{FeelVOS} & ResNet-$50$ & &  {\color{gray} Offline} & - & $26.9$ & $42.0$ & $29.7$ & $29.9$ & $33.4$ \\
  SeqTracker \cite{VIS} & ResNet-$50$ & & {\color{gray} Offline} & - & $27.5$ & $45.7$ & $28.7$ & $29.7$ & $32.5$ \\
  MaskTrack R-CNN \cite{VIS} & ResNet-$50$ & & Online & $32.8$ & $30.3$ & $51.1$ & $32.6$ & $31.0$ & $35.5$ \\
  MaskProp \cite{MaskPropagation} & ResNet-$50$ & \checkmark\checkmark & {\color{gray} Offline} & $< 6.2^{\dag}$ & $40.0$ & - & $42.9$ & - & - \\
  SipMask-VIS \cite{SipMask} & ResNet-$50$ & & Online & $34.1$ & $32.5$ & $53.0$ & $33.3$ & $33.5$ & $38.9$ \\
  SipMask-VIS \cite{SipMask} & ResNet-$50$ & \checkmark & Online & $34.1$ & $33.7$ & $54.1$ & $35.8$ & $35.4$ & $40.1$ \\
  STEm-Seg \cite{STEm-Seg} & ResNet-$50$ & \checkmark\checkmark & {\color{gray} Near Online} & $4.4$ & $30.6$ & $50.7$ & $33.5$ & $31.6$ & $37.1$ \\
  Johnander \etal \cite{VIS_RGNN} & ResNet-$50$ & \checkmark\checkmark & Online & $\sim 30$ & $35.3$ & - & - & - & - \\
  CompFeat \cite{CompFeat} & ResNet-$50$ & \checkmark\checkmark & Online & $< 32.8$ & $35.3$ & $56.0$ & $38.6$ & $33.1$ & $40.3$ \\
  VisTR \cite{VisTR} & ResNet-$50$ & & {\color{gray} Offline} & $30.0$ & $34.4$ & $55.7$ & $36.5$ & $33.5$ & $38.9$ \\
  \textbf{CrossVIS} & ResNet-$50$ & & Online & $39.8$ & $34.8$ & $54.6$ & $37.9$ & $34.0$ & $39.0$ \\
  \textbf{CrossVIS} & ResNet-$50$ & \checkmark & Online & $39.8$ & $36.3$ & $56.8$ & $38.9$ & $35.6$ & $40.7$ \\
  \textbf{CrossVIS-Lite} & DLA-$34$ & & Online & $48.5$ & $33.0$ & $52.7$ & $35.0$ & $33.9$ & $39.5$ \\
  \textbf{CrossVIS-Lite} & DLA-$34$ & \checkmark & Online & $48.5$ & $36.2$ & $56.7$ & $38.4$ & $35.1$ & $42.0$ \\
  \hline
  MaskTrack R-CNN \cite{VIS} & ResNet-$101$ & & Online & $28.6$ & $31.9$ & $53.7$ & $32.3$ & $32.5$ & $37.7$ \\
  MaskProp \cite{MaskPropagation} & ResNet-$101$ & \checkmark\checkmark & {\color{gray} Offline} & $< 5.6^{\dag}$ & $42.5$ & - & $45.6$ & - & - \\
  STEm-Seg \cite{STEm-Seg} 
  & ResNet-$101$ 
  & \checkmark\checkmark 
  & {\color{gray} Near Online} 
  & $2.1$
  & $34.6$ & $55.8$ & $37.9$ & $34.4$ & $41.6$ \\
  VisTR \cite{VisTR} 
  & ResNet-$101$ 
  & 
  & {\color{gray} Offline} 
  & $27.7$ & $35.3$ & $57.0$ & $36.2$ & $34.3$ & $40.4$ \\
  \textbf{CrossVIS} 
  & ResNet-$101$ 
  & 
  & Online 
  & $35.6$ & $36.6$ & $57.3$ & $39.7$ & $36.0$ & $42.0$ \\
 \hline
 
 \hline
 \end{tabular}
 \end{small}
 \medskip
 \caption{Comparisons with some state-of-the-art VIS models on \textbf{YouTube-VIS-2019} $\mathtt{val}$ set. 
 The compared methods are listed roughly in the temporal order.
 ``\checkmark" under ``Aug." indicates using multi-scale input frames during training.
 ``\checkmark\checkmark" indicates using stronger data augmentation (\eg, random crop, higher resolution input, \etc) \cite{MaskPropagation, VIS_RGNN} or additional data \cite{STEm-Seg, CompFeat, VIS_RGNN}. 
 The FPS with superscript ``${\dag}$" is not reported in \cite{MaskPropagation} and is estimated using its utilized components \cite{HTC, ResNet, ResNeXt, STSN}. For the definition of online and offline, we follow \cite{TSM, MOT_Review}.}
 \label{tab: main results}
\end{table*}

  
  

\section{Experiments}

\subsection{Dataset}

We evaluate the proposed CrossVIS on three challenging video instance segmentation benchmarks, \ie, YouTube-VIS-2019 \cite{VIS}, OVIS \cite{OVIS} and YouTube-VIS-2021 \cite{YouTube-VIS-2021}.

\begin{itemize}
    \item 
	\textbf{YouTube-VIS-2019} is the first dataset for video instance segmentation, which has a $40$-category label set, $4,883$ unique video instances and $131k$ high-quality manual annotations.
	There are $2,238$ training videos, $302$ validation videos, and $343$ test videos in it. 
	\item 
	\textbf{OVIS} dataset is a recently proposed very challenging VIS dataset with the philosophy of perceiving \textit{object occlusions} in videos, which could reveal the complexity and the diversity of real-world scenes. 
	OVIS consists of $296k$ high-quality instance masks (about $2 \times$ of YouTube-VIS-2019) and $5.80$ instance per video (about $3.4 \times$ of YouTube-VIS-2019) from $25$ semantic categories, where object occlusions usually occur.
	There are $607$ training videos, $140$ validation videos, and $154$ test videos in this dataset. 
	\item 
	\textbf{YouTube-VIS-2021} dataset is an improved and augmented version of YouTube-VIS-2019 dataset, which has $8,171$ unique video instances and $232k$ high-quality manual annotations (about $2 \times$ of YouTube-VIS-2019).
	There are $2,985$ training videos, $421$ validation videos, and $453$ test videos in this dataset.
	\end{itemize}

	Unless specified, AP and AR in this paper refer to the average precision and average recall defined in \cite{VIS}.
	Following previous works \cite{VIS, SipMask, MaskPropagation, STEm-Seg}, we report our results on the validation set to evaluate the effectiveness of the proposed method.

\subsection{Implementation Details}
    \noindent \textbf{Basic Setup.} \ 
    Similar to the setup of \cite{VIS, SipMask}, we initialize CrossVIS with corresponding CondInst instance segmentation model \cite{CondInst, ResNet, DLA, FPN} pre-trained on COCO $\mathtt{train2017}$ \cite{COCO} with $1 \times$ schedule.
    Then we train the CrossVIS on VIS datasets with $1 \times$ schedule. 
    The pre-train procedure on COCO follows $\mathtt{Detectron2}$ \cite{Detectron2} and $\mathtt{AdelaiDet}$ \cite{AdelaiDet}.
    $1 \times$ schedule on VIS datasets refers to $12$ epoch \cite{VIS}.
    The learning rate is set to $0.005$ initially following SipMask-VIS \cite{SipMask} and reduced by a factor of $10$ at epoch $9$ and $11$.
	Most FPS data\footnote{Please note that some FPS data (\eg, \cite{VisTR, VIS_RGNN}) in Tab.~\ref{tab: main results} is measured using V$100$, which is \textit{slightly faster} than measured using $2080$ Ti.} is measured with an RTX $2080$ Ti GPU including the pre- and post-processing steps.
	We report the results using the median of 5 runs \cite{PointRend, LVIS} due to the variance inherent in VIS models and datasets.

	\noindent \textbf{Input Frame Size.} \ 
	For single-scale training, we resize the frame to $360 \times 640$. 
	For multi-scale training, we follow the setting in SipMask-VIS.
	During inference, we resize the frame to $360 \times 640$.
	
    
    

    \noindent \textbf{Main Results.} \ 
    For our main results, we evaluate the proposed CrossVIS on YouTube-VIS-2019, OVIS and YouTube-VIS-2021 datasets, respectively.
	
	\noindent \textbf{Ablation Study.} \ 
	Our ablation study is conducted on the YouTube-VIS-2019 dataset using models with ResNet-$50$-FPN \cite{ResNet, FPN} backbone.

\subsection{Main Results}

\noindent \textbf{Main Results on YouTube-VIS-2019 Dataset.} \ 
    We compare CrossVIS against some state-of-the-art methods in Tab.~\ref{tab: main results}. 
    The comparison is performed in terms of both accuracy and speed.
    $(1)$ When using the single-scale training strategy, CrossVIS achieves $34.8$ AP using ResNet-$50$ and $36.6$ AP using ResNet-$101$, which is the best among all the online and near online methods in Tab.~\ref{tab: main results}.
    CrossVIS also outperforms the recently proposed offline method VisTR.
    $(2)$ When using the multi-scale training strategy, CrossVIS achieves $36.3$ AP and $39.8$ FPS with ResNet-$50$, which outperforms SipMask-VIS, STEm-Seg and VisTR with the stronger ResNet-$101$ backbone.
    $(3)$ Moreover, CrossVIS achieves the best speed-accuracy trade-off among all VIS approaches in Tab.~\ref{tab: main results}.
    We also present a more efficient CrossVIS-Lite model with DLA-$34$ backbone, achieving $36.2$ AP and $48.5$ FPS, which shows a decent trade-off between latency and accuracy. 
    
    MaskProp \cite{MaskPropagation} is a state-of-the-art offline VIS approach that proposes a novel mask propagation mechanism in combination with Spatiotemporal Sampling Network \cite{STSN}, Hybrid Task Cascade mask head \cite{HTC}, High-Resolution Mask Refinement post-processing, longer training schedule and stronger data argumentation.
    MaskProp can achieve very high accuracy but suffer from low inference speed so it is far from real-time applications and online scenarios.
    Meanwhile, CrossVIS is designed to be an efficient online VIS model and focusing more on the speed-accuracy trade-off.
    Overall, the experiment results demonstrate the effectiveness of the proposed approach. 

\begin{table}
 \setlength{\tabcolsep}{5.1 pt}
 \begin{small}
 \begin{tabular}{l|c|ccc}
 \hline
  
 \hline
  Methods 
  & Calibration \cite{OVIS}
  & AP 
  & AP$_{\mathtt{50}}$ 
  & AP$_{\mathtt{75}}$ 
  \\
  \hline
  \hline
  SipMask-VIS & & $10.3$ & $25.4$ & $7.8$ \\
  MaskTrack R-CNN & & $10.9$ & $26.0$ & $8.1$ \\
  STEm-Seg & & $13.8$ & $32.1$ & $11.9$ \\
  \textbf{CrossVIS} & & ${14.9}$ & ${32.7}$ & ${12.1}$ \\
  \hline
  SipMask-VIS & \checkmark & $14.3$ & $29.9$ & $12.5$ \\
  MaskTrack R-CNN & \checkmark & $15.4$ & $33.9$ & $13.1$ \\
  \textbf{CrossVIS} & \checkmark & ${18.1}$ & ${35.5}$ & ${16.9}$ \\
 \hline
  
 \hline
 \end{tabular}
 \end{small}
 \medskip
 \caption{Comparisons with some VIS models on the recently proposed very challenging \textbf{OVIS} $\mathtt{val}$ set. We use ResNet-$50$ backbone and $1 \times$ schedule for all experiments.}
 \label{tab: ovis results.}
\end{table}

\noindent \textbf{Main Results on OVIS Dataset.} \
    OVIS is a much more challenging VIS benchmark than YouTube-VIS-2019 and all methods encounter a large performance degradation on this dataset.
    CrossVIS achieves $14.9$ AP and $18.1$ AP with the temporal feature calibration module proposed in \cite{OVIS}, surpassing all methods investigated in \cite{OVIS} under the same experimental conditions.
    We hope CrossVIS can serve as a strong baseline for this new and challenging benchmark.

\begin{table}[t]
 \setlength{\tabcolsep}{4.2 pt}
 \begin{small}
 \begin{tabular}{l|c|ccccc}
  \hline
    
  \hline
  Methods 
  & Aug.  
  & AP 
  & AP$_{\mathtt{50}}$ 
  & AP$_{\mathtt{75}}$ 
  & AR$_{\mathtt{1}}$ 
  & AR$_{\mathtt{10}}$ 
  \\
 \hline
 \hline
  MaskTrack R-CNN & 
  & $28.6$ & $48.9$ & $29.6$ & $26.5$ & $33.8$ \\
  SipMask-VIS & \checkmark
  & $31.7$ & $52.5$ & $34.0$ & $30.8$ & $37.8$ \\
  \textbf{CrossVIS} &
  & $33.3$ & $53.8$ & $37.0$ & $30.1$ & $37.6$ \\
  \textbf{CrossVIS} & \checkmark
  & $34.2$ & $54.4$ & $37.9$ & $30.4$ & $38.2$ \\
 \hline
    
 \hline
 \end{tabular}
 \end{small}
 \medskip
 \caption{Comparisons with some VIS models on the recently proposed \textbf{YouTube-VIS-2021} $\mathtt{val}$ set. We use ResNet-$50$ backbone and $1 \times$ schedule for all experiments.}
 \label{tab: youtube-vis-2021 results.}
\end{table}

\noindent \textbf{Main Results on YouTube-VIS-2021 Dataset.} \ 
    YouTube-VIS-2021 dataset\footnote{This dataset is released in February, 2021. To our knowledge, there is no publicly available baseline result on this dataset by the time we submit this preprint to arXiv.} is an improved and augmented version of YouTube-VIS-2019 dataset. 
    We evaluate the recently proposed MaskTrack R-CNN and SipMask-VIS on this dataset using official implementation for comparison.
    As shown in Tab.~\ref{tab: youtube-vis-2021 results.}, CrossVIS surpasses MaskTrack R-CNN and SipMask-VIS by a large margin.
    We hope CrossVIS can serve as a strong baseline for this new and challenging benchmark.


\subsection{Ablation Study}
\label{ablation}

\noindent \textbf{Does Better VIS Results Simply Come from Better Still-image Instance Segmentation Models?} \
    The answer is no.
    We prove this in Tab.~\ref{tab: coco ap controlled.}:
    $(1)$ 
    Compared with MaskTrack R-CNN using ResNet-$101$ backbone, CrossVIS is $0.2$ AP$^{\mathtt{COCO}}_\mathtt{{mask}}$ lower, which 
    indicates that our pre-trained model is relatively weaker in terms of still-image instance segmentation on COCO.
    But for the VIS task, our model is $2.9$ AP$^{\mathtt{VIS}}$ higher.
    $(2)$ 
    We implement a VIS baseline called CondInst-VIS which replaces the Mask R-CNN part in MaskTrack R-CNN by CondInst.
    Therefore the only differences between CrossVIS and CondInst-VIS are the proposed crossover learning scheme and global balanced instance embedding branch.
    Compared with CondInst-VIS with ResNet-$50$ and MaskTrack R-CNN with ResNet-$101$, we conclude that they achieve similar AP$^{\mathtt{VIS}}$ under similar AP$^{\mathtt{COCO}}_\mathtt{{mask}}$.
    Meanwhile, CrossVIS is $2.7$ AP$^{\mathtt{VIS}}$ better than CondInst-VIS under the same AP$^{\mathtt{COCO}}_\mathtt{{mask}}$.
    The above two observations prove that the improvement in AP$^{\mathtt{VIS}}$ mainly comes from the proposed two modules instead of better pre-trained models or baseline.
    
\begin{table}
  \begin{small}
  \setlength{\tabcolsep}{4.86 pt}
  \begin{tabular}{l|l|c|cc}
  \hline
  
  \hline
  Method
  & Backbone 
  & Sched. 
  & AP$^{\mathtt{VIS}}$ 
  & AP$^{\mathtt{COCO}}_\mathtt{{mask}}$ \\
  \hline
  \hline
  MaskTrack R-CNN & ResNet-$50$ & \multirow{4}{*}{$1 \times$} & $30.3$ & $34.7$ \\
  MaskTrack R-CNN & ResNet-$101$ & & $31.9$ & $\mathbf{35.9}$ \\
  CondInst-VIS & ResNet-$50$ & & $32.1$ & $35.7$ \\
  \textbf{CrossVIS} & ResNet-$50$ & & $\mathbf{34.8}$ & $35.7$ \\
  \hline
  
  \hline
 \end{tabular}
 \end{small}
 \smallskip
 \caption{Comparisons between CrossVIS and other baselines in terms of both AP$^{\mathtt{VIS}}$ and AP$^{\mathtt{COCO}}_\mathtt{{mask}}$ on YouTube-VIS-2019 $\mathtt{val}$ set.
 }
 \label{tab: coco ap controlled.}
\end{table}

\begin{table}[t]
 \begin{small}
 \setlength{\tabcolsep}{4.8 pt}
 \begin{tabular}{l|l|l|c}
  \hline
  
  \hline
  Method 
  & Backbone 
  & FPS ($ 360 \times 640 $)
  & AP$^{\mathtt{VIS}}$  \\
  \hline
  \hline
  Mask R-CNN \cite{MaskRCNN} 
  & \multirow{2}{*}{ResNet-$50$} 
  & \ \ \ \ $41.6$ & - \\
  
  CondInst \cite{CondInst} 
  & 
  & \ \ \ \ $42.8 \ (+ 1.2)$ & - \\
  \hline
  MaskTrack R-CNN 
  & \multirow{2}{*}{ResNet-$50$}
  & \ \ \ \ $32.8$ & $30.3$ \\
  
  \textbf{CrossVIS} 
  & 
  & \ \ \ \ $39.8 \ (+ 7.0)$ & $34.8$ \\
 \hline
 
 \hline
 \end{tabular}
 \end{small}
 \smallskip
 \caption{Efficiency comparisons on YouTube-VIS-2019 $\mathtt{val}$ set.}
 \label{tab: latency-accuracy trade-off}
\end{table}

\begin{table*}[t]
	\begin{small}
	\setlength{\tabcolsep}{10.37 pt}
	\begin{tabular}{c|l|l|l|l|l|l|l}
		\hline
		
		\hline
		Crossover & $T = 1$ & $T = 3$ & $T = 5$ & $T = 10$ & $T = 15$ & $T = 20$ & $T = \infty$ \\
		\hline
		\hline
		& $33.1$ & $33.4$ & $33.5$ & $33.5$ & $33.6$ & $33.4$ & $33.5$ \\
		\checkmark & $33.6_{\uparrow(0.5)}$ & $34.2_{\uparrow(+0.8)}$ & $34.6_{\uparrow(+1.1)}$ & $34.6_{\uparrow(+1.1)}$ & $34.3_{\uparrow(+0.7)}$ & $\mathbf{34.8}_{\uparrow(+1.4)}$ & $\mathbf{34.8}_{\uparrow(+1.3)}$ \\
		\hline
		
		\hline
	\end{tabular}
	\end{small}
	\smallskip
	\caption{Effect of crossover learning and sample time interval on AP. We randomly sample two frames at time $t$ and $t + \delta$ respectively, where the sample time interval $\delta \in [-T, T]$. The ``$\uparrow$" indicates the AP improvement of the model with crossover learning compared to the model without crossover learning under the same $T$.} 
	\label{tab: ablation for crossover learning}
\end{table*}

\begin{table}[t]
 \begin{small}
 \setlength{\tabcolsep}{8.0 pt}
 \begin{tabular}{ll|ccc}
 \hline
 
 \hline
  Embedding 
  & Loss 
  & AP $\pm \ \sigma_{\text{AP}}$
  & AP$_{\mathtt{50}}$ 
  & AP$_{\mathtt{75}}$
  \\
  \hline
  \hline
  Pair-wise & $\mathcal{L}_{CE}$ & $33.1 \pm 0.78$ & $51.9$ & $34.9$ \\
  Pair-wise & $\mathcal{L}_{Focal}$ & $33.3 \pm 0.72$ & $52.1$ & $35.0$ \\
  Global & $\mathcal{L}_{CE}$ & $33.4 \pm 0.27$ & $53.9$ & $35.7$ \\
  Global & $\mathcal{L}_{Focal}$ & $\mathbf{34.8} \pm \mathbf{0.25}$ & $\mathbf{54.6}$ & $\mathbf{37.9}$ \\
 \hline
 
 \hline
 \end{tabular}
 \end{small}
 \smallskip
 \caption{Study of instance association embeddings. To quantitate the fluctuation in results, we conduct $5$ independent experiments for each configuration. We report the AP using the \textit{median} of $5$ runs. $\sigma_{\text{AP}}$ denotes the \textit{standard deviation} of $5$ runs. 
 }  
 \label{tab: ablation for instance association embedding}
\end{table}

\noindent \textbf{Does the Efficiency of CrossVIS Simply Come from the Efficiency of CondInst?} \ 
    The answer is no. 
    We prove this in Tab.~\ref{tab: latency-accuracy trade-off}.
    In terms of the inference speed, CondInst is only $1.2$ FPS faster than Mask R-CNN in instance segmentation task (similar conclusions are also reported in \cite{CondInst}). 
    Meanwhile, CrossVIS is $7.0$ FPS faster than MaskTrack R-CNN in VIS task.
    This is mainly because: 
    $(1)$ crossover learning adds no extra parameters and can bring cost-free improvement during inference. 
    $(2)$ The global balanced embedding branch adopts a lightweight fully convolutional design compared to the fully connected design in MaskTrack R-CNN. 
    Therefore the efficiency of CrossVIS mainly comes from the efficient design of crossover learning and global balanced embedding.

\noindent \textbf{Crossover Learning.} \ 
	Here we investigate the effectiveness of the proposed crossover learning scheme in Sec.~\ref{sec: crossover}. 
	During training, we randomly sample frame pairs with a sample time interval $\delta \in [-T, T]$.
	The results are shown in Tab.~\ref{tab: ablation for crossover learning}. 
	We conclude that:
	$(1)$ When the sample time interval $\delta$ is small, \textit{e.g.}, $\delta = [-1, 1]$, the crossover learning brings moderate improvement compared to the baseline. 
	This makes sense because the sampled two frames are quite similar to each other when the $\delta$ is small.
	Under this circumstance, the crossover learning degenerates to na\"ive still-image training.
	$(2)$ When the sample time interval $\delta$ becomes larger, the scene and context become different and diverse between two frames in the sampled frame pair. 
	The baseline without crossover learning cannot explicitly utilize the cross-frame information therefore only has limited improvement.
	However, crossover learning can benefit significantly from the larger $\delta$ and achieves up to $1.4$ AP improvement compared to the baseline.
	$(3)$ The proposed crossover learning scheme is quite insensitive to the variations of $T$.
	Overall, models trained with crossover scheme are $\sim 1$ AP higher than baselines under a wide range of time intervals, \textit{i.e.}, from $T = 3$ to $T = \infty$ as shown in Tab.~\ref{tab: ablation for crossover learning}. 
	
	These results prove the analysis in Sec.~\ref{sec: crossover} that the crossover scheme can leverage the rich contextual information across video frames to get a more accurate and robust instance representation.

\noindent \textbf{Instance Association Embeddings.} \ 
We study the instance association embeddings in Tab.~\ref{tab: ablation for instance association embedding}. 
As expected in Sec.~\ref{sec: instance embeddings}, 
$(1)$ In terms of AP, the effect from ``global" (using learnable $\{\mathbf{w}_M\}$ instead of sampled $\{\mathbf{e}_N\}$) and ``balanced" (using $\mathcal{L}_{Focal}$ instead of $\mathcal{L}_{CE}$) are \textit{equally important and interdependent}:
Using $\mathcal{L}_{Focal}$ instead of $\mathcal{L}_{CE}$ for pair-wise embedding can only bring $0.2$ AP improvement, for large pos-neg imbalance do not exist in the pair-wise scheme. 
Using global instead of pair-wise embedding optimized by $\mathcal{L}_{CE}$ can only bring $0.3$ AP improvement, for there exists a large pos-neg imbalance issue.
But together, global and balanced embedding can bring $1.7$ AP improvement. 
So global and balanced are both \textit{indispensable} for good performance.
$(2)$ In terms of AP fluctuation, using the global embedding has a much smaller standard deviation $\sigma_{\text{AP}}$ than the pair-wise embedding regardless of the loss function, which indicates that the global embedding can produce a more definite convergence status and more stable results.

\begin{table}[t]
  \center
  \begin{small}
  \setlength{\tabcolsep}{15.0 pt}
  \begin{tabular}{ccc|c}
  \hline
  
  \hline
    Baseline
  & COL
  & GBE
  & AP 
  \\
  \hline
  \hline
  \checkmark & & & $32.1$ \\
  \checkmark & \checkmark & & $33.1$ \\
  \checkmark & & \checkmark & $33.5$ \\
  \checkmark & \checkmark & \checkmark & $34.8$ \\
  \hline
  
  \hline
 \end{tabular}
 \end{small}
 \smallskip
 \caption{Impact of integrating \textbf{C}ross\textbf{O}ver \textbf{L}earning (COL) and \textbf{G}lobal \textbf{B}alanced \textbf{E}mbedding (GBE) into CondInst-VIS baseline.}
 \label{tab: analysis of each component.}
\end{table}

\noindent \textbf{Component-wise Analysis.} \ 
We investigate the effects of crossover learning and global balanced embedding individually and simultaneously in Tab.~\ref{tab: analysis of each component.}. 
Using crossover learning and global balanced embedding individually can bring $1.0$ AP and $1.4$ AP improvement, respectively.
In terms of AP, global balanced embedding is slightly higher. 
Meanwhile, crossover learning adapts CondInst naturally for VIS task during training and brings cost-free improvement during inference.
Together, the two components bring $2.7$ AP improvement, which is larger than $1.0 + 1.4$ AP when used solely.
Therefore the proposed two components are fully compatible with each other. 
They show synergy and their improvements are complementary.

\begin{figure*}[t!]
    \centering
    \includegraphics[width=2.0\columnwidth]{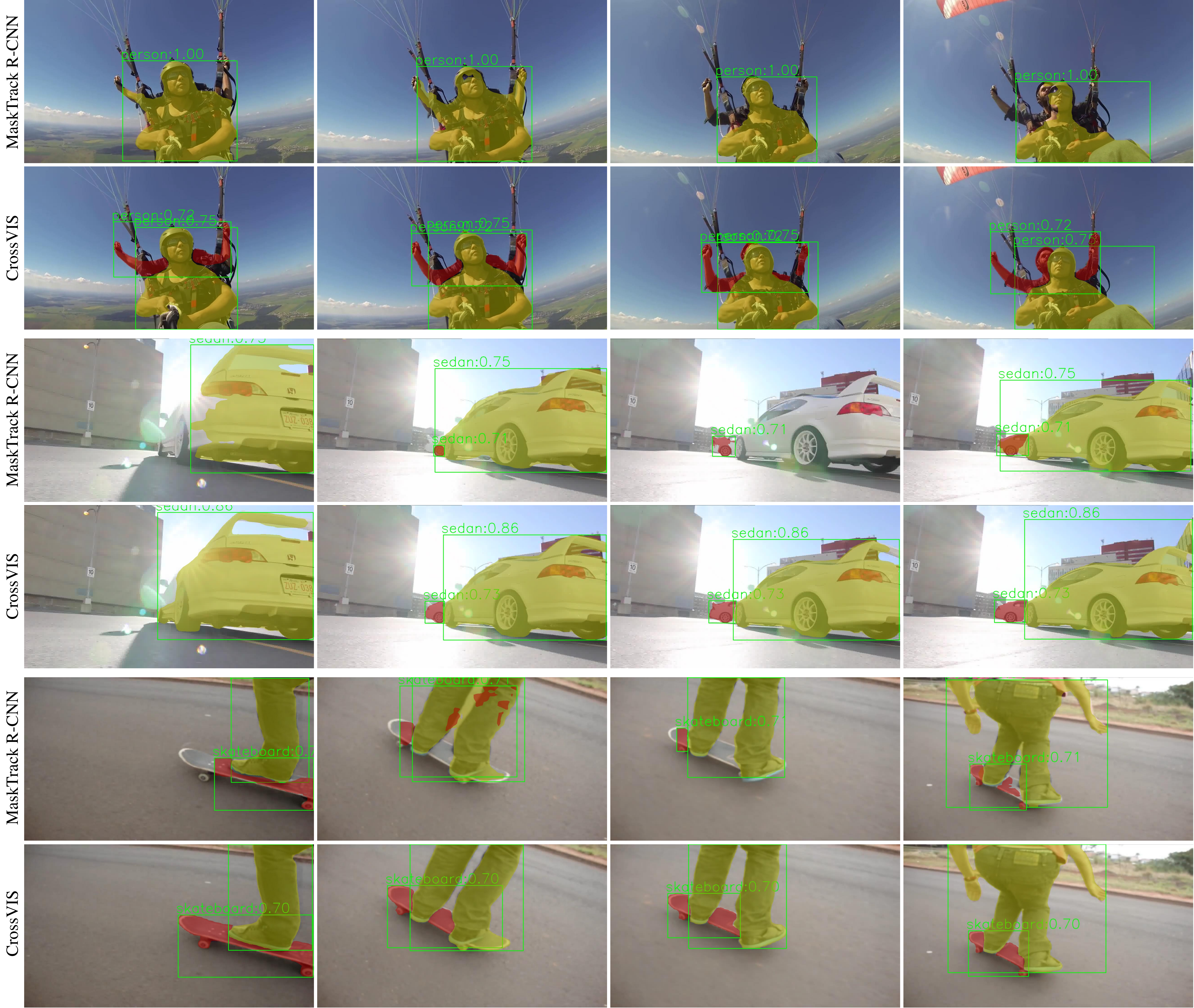}
    \caption{We compare some qualitative VIS results obtained by CrossVIS and MaskTrack R-CNN \cite{VIS}. Different color masks belong to different instances.}
    \label{fig: additional qualitative results 2}
\end{figure*}

\begin{figure*}[t!]
    \centering
    \includegraphics[width=2.0\columnwidth]{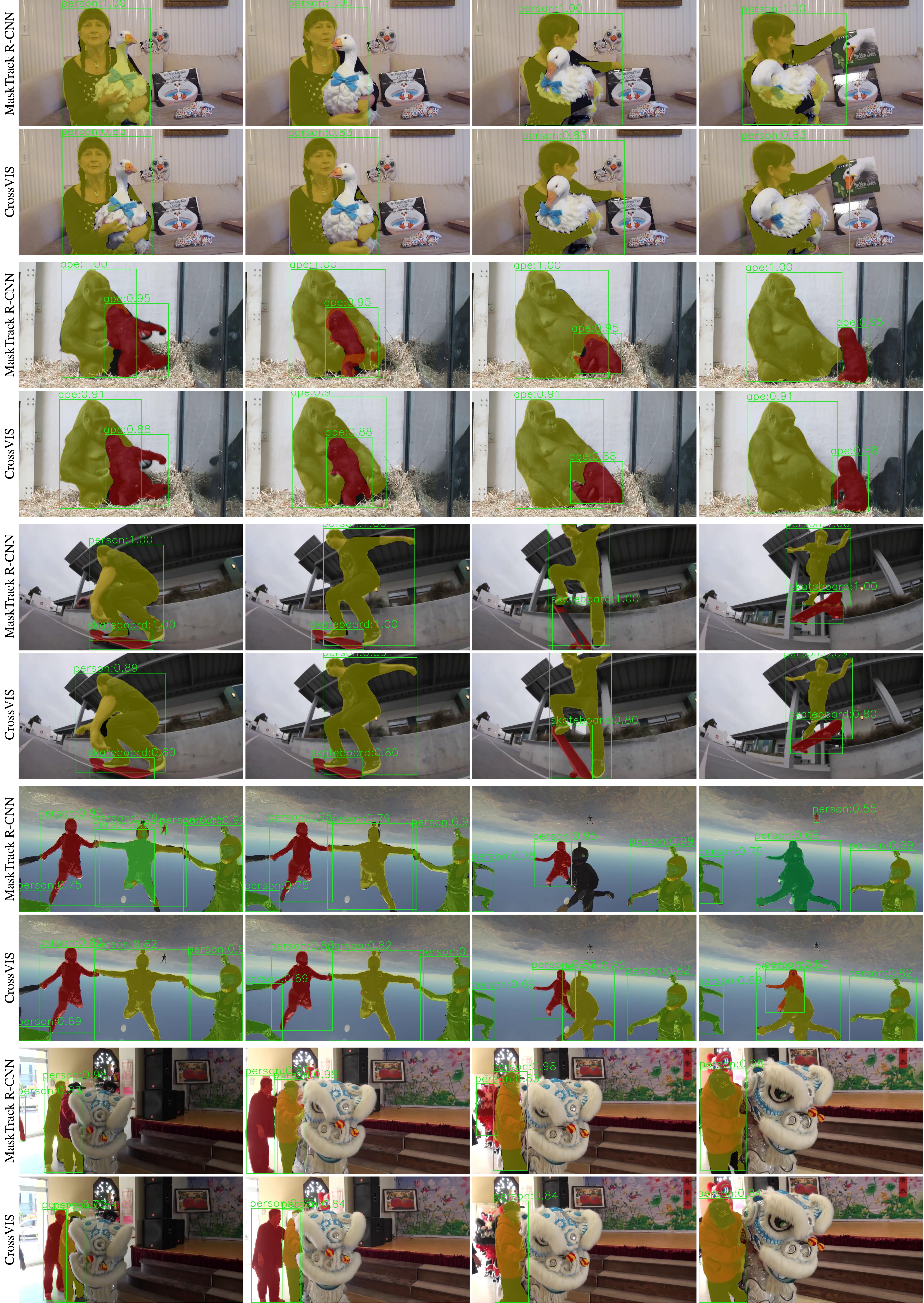}
    \caption{We compare some qualitative VIS results obtained by CrossVIS and MaskTrack R-CNN \cite{VIS}. Different color masks belong to different instances.}
    \label{fig: additional qualitative results 1}
\end{figure*}

\section{Conclusion}
	In this paper, we introduce a novel VIS solution coined as CrossVIS, which performs the best among all online video instance segmentation methods in three challenging VIS benchmarks.
	Moreover, CrossVIS strikes a decent trade-off between latency and accuracy. 
	We also show that the accuracy and efficiency of CrossVIS are not simply come from the instance segmentation framework but stems from the proposed design.
	Extensive study proves that crossover learning can bring cost-free improvement during inference, while the lightweight global balanced embedding can help stabilize the model performance. 
	We believe that the proposed approach can serve as a strong baseline for further research on the VIS, and sheds light on other video analysis and video understanding tasks.


\section*{Appendix}

\noindent \textbf{Qualitative Results.} \
	We compare some qualitative VIS results obtained by CrossVIS and MaskTrack R-CNN in Fig.~\ref{fig: additional qualitative results 2} and Fig.~\ref{fig: additional qualitative results 1}.
	Our CrossVIS segments and tracks object instances more robustly even if they are occluded or overlap with each other.

\clearpage

{\small
\bibliographystyle{ieee_fullname}
\bibliography{crossvis_arxiv_v1}
}

\end{document}